\documentclass[preprint,12pt,3p,authoryear]{elsarticle}
\usepackage{lineno,hyperref}
\usepackage[usenames, dvipsnames]{xcolor}
\usepackage[title]{appendix}
\usepackage{soul}
\usepackage{amsmath}
\usepackage{float}
\usepackage{subcaption}
\usepackage{adjustbox}
\modulolinenumbers[5]
\usepackage{tabularx,booktabs}
\definecolor{sand}{RGB}{255,255,187}
\definecolor{shale}{RGB}{142,142,139}
\usepackage{amssymb}
\usepackage{ulem}
\usepackage{multirow}
\usepackage{amsmath}

\journal{Journal of Petroleum Science and Engineering}

\author[label1,label2, label4,label5,label8]{Ekaterina Gurina\corref{cor1}}
\cortext[cor1]{Corresponding author}
\ead{Ekaterina.Gurina@skoltech.ru}

\author[label1,label2,label4,label5,label8]{Nikita Klyuchnikov}
\author[label1,label2,label9,label10]{Ksenia Antipova}
\author[label1,label2,label9,label10]{Dmitry Koroteev}

\address[label1]{Skolkovo Institute of Science and Technology (Skoltech), 121205, Moscow, Russia}
\address[label2]{Digital Petroleum, Skolkovo Innovation Center, 121205,  Moscow,  Russia}

\fntext[label4]{Conceptualization}
\fntext[label5]{Implementation methods}
\fntext[label8]{Drafting the manuscript}
\fntext[label9]{Revising the manuscript}
\fntext[label10]{Project administration}

\begin{document}
\begin{frontmatter}

\title{Making the black-box brighter: interpreting machine learning algorithm for forecasting  drilling accidents}

\begin{abstract}
We present an approach for interpreting a black-box alarming system for forecasting accidents and anomalies during the drilling of oil and gas wells. The interpretation methodology aims to explain the local behavior of the accident predictive model to drilling engineers. The explanatory model uses Shapley additive explanations analysis of features, obtained through Bag-of-features representation of telemetry logs used during the drilling accident forecasting phase. Validation shows that the explanatory model has 15\% precision at 70\% recall, and overcomes the metric values of a random baseline and multi-head attention neural network. These results justify that the developed explanatory model is better aligned with explanations of drilling engineers, than the state-of-the-art method. The joint performance of explanatory and Bag-of-features models allows drilling engineers to understand the logic behind the system decisions at the particular moment, pay attention to highlighted telemetry regions, and correspondingly, increase the trust level in the accident forecasting alarms.

\end{abstract}


\end{frontmatter}

\section{Introduction}
\label{sec:intro}
\modulolinenumbers[1]
Artificial intelligence (AI) is rapidly spreading over different areas and industries, creating significant potential for growth and possible optimization of existing solutions \citep{intelligence2019wipo, bughin2018notes}. AI is usually represented by machine learning (ML) and deep learning (DL) models, adopted and trained for the particular problem, that provide quantitative solutions. In the oil and gas industry, namely, in the upstream section, AI is used for such problems as a prediction of different reservoir properties \citep{erofeev2019prediction} and logs \citep{rostamian2022evaluation},well placement optimization \citep{rostamian2019development, rostamian2019non, rostamian2017well}, lithology classification \citep{klyuchnikov2019data}, modeling during hydraulic fracturing \citep{makhotin2019gradient}, forecasting of material properties and cracks in the drill strings \citep{sobhaniaragh2021towards, sobhaniaragh2020hydrogen, mirseyed2022novel, afzalimir2020evaluation}, others. 

In papers \citep{gurina2022forecasting, antipova2019data} authors consider the problem of minimizing non-productive time (NPT) during the well construction process using AI techniques. At each time moment, the proposed ML model returns the probability of whether the corresponding time interval contains anomaly behavior that can lead to the specific type of accident or not. The model is based on the Bag-of-features representation of telemetry logs and works in real-time. Nowadays, it is the state-of-art (SOTA) method for drilling accident forecasting problem that is successfully used during drilling in real oilfields \citep{novatek}. Based on the results obtained for the last year, the company has $85$ drilling accidents, among which the predictive model was able to forecast $60$ cases, and helped to reduce NPT by $15$ percent what is demonstrated in reference \citep{novatek}. 

Since drilling of oil and gas wells is a complex process, drilling engineers usually require not only the probability of an accident but also the explanation of the intelligence behind the predictive model output to be sure in making the decision based on it. Usage of an unknown black-box model is often accompanied by fears of incorrect recommendations and low trust in the model forecast. Such behavior can slow down or even stop the model's integration process in everyday use. Interpretability of model behavior helps to overcome some of these issues. Besides, understating model logic is also crucial for machine learning engineers, who can improve the model through feature engineering, parameter tuning, or even replacing the model with a different one, if it learns irrelevant patterns during training and, consequently, has overfitting problems.  

 Interpretability is a critical property that has been investigated in various research areas, from healthcare to the mining industry \citep{gilpin2018explaining}. The first issue that researchers faced dealing with interpretability is the lack of problem formalization. Such problem was raised in papers \citep{kaur2020interpreting, nori2019interpretml} where authors discuss different aspects of interpretability concept and tried to formulate general frameworks that can be applied in most studies. There have been many attempts to introduce the interpretability concept to this date, and it has been studied from different points of view in paper \citep{carvalho2019machine}.Authors summarize that there are two possible ways to interpret the model: use interpretable-by-nature models or create explanation methods that can be applied to existing black-box models. In the end, they concluded, that interpretability is a domain-specific notion, and it is necessary to take into consideration the application domain where the interpretability is applied to. In this study, we are not focused on the question of formulating the concept of interpretability. Thus, we will rely on the definition introduced in the article \citep{ribeiro2016model} since it is the most used one in the literature. According to the article, interpretability is a rationale for why a model's prediction was made, which is based on the relative influence of the features used during the inference, and which should be provided using textual or visual components of the data.

The main contributions of this paper are:
\begin{itemize}
  \item we upgrade the black-box accidents Bag-of-features model to the interpretable one based on the Shapley method;
  \item we validate the quality of model explanations against experts' annotations and compare the performance with the SOTA method based on neural attention mechanism;
  \item we verify the consistency of explanations using visual investigation of the model's behavior.
\end{itemize}

The paper is organized as follows: Section \ref{sec:state-of-art} contains a brief introduction to the existing solution that provides interpretability for the black-box model and methods used for interpretability quality assessments. In Section \ref{sec:methodology} we presented the methodology of the current study, while section \ref{sec:results} provides the results obtained during different experiments. Finally, Sect. \ref{sec:conclusion} presents the conclusions of this study.

\section{Related work}
\label{sec:state-of-art}
Since the current study is mostly aimed at studying the interpretability techniques for black-box machine learning models, in this section we describe methods that are used nowadays to interpret forecasting black-box models and assess the quality of obtained interpretability. The full literature review related to the forecasting of drilling accidents can be found in papers \citep{gurina2021machine, gurina2022forecasting, gurina2022application}, while here we mostly focus on the review of the system that provides forecast and explanation why it was made.

\subsection{Methods for predicting drilling accidents}

Nowadays, there are several existing solutions to forecast drilling accidents. In papers \citep{borozdin2020drilling, aljubran2021deep, gurina2022application} authors use neural networks or machine learning algorithms as a model to forecast drilling accidents. The idea is to train a model using different drilling parameters for several time intervals in the past, which allows using these solutions in real-time as input features. As output model returns the probability of whether the current interval is similar to accident patterns fitted to the model during the training phase or not. Since it is the black-box solution, such type of models does not explain their behavior.

Another group of solutions is based on the analogs search of the current situation from the database of past events, that happened in different oil and gas wells. In papers \citep{ferreira2015automated, gurina2020application}, authors use black-box models with different feature engineering approaches to forecasting the most common types of drilling accidents. Similar to the previous group of methods, the system returns the interval from the real-case database, which is the most similar to the current situation according to formal criteria. Visual comparison of both intervals by the drilling engineer may give some insights into system logic, however, does not guarantees it. 

The last group of methods includes solutions that use mathematical modeling inside the decision module. For example, in paper \citep{da2020six}, the authors use an online numerical simulation of a possible drilling event based on the solution of a system of partial differential equations. The system returns the predicted values of the main drilling parameters. Next, the drilling engineer independently assesses whether the deviation of real data from the obtained one can lead to an accident or not.

Based on research results obtained through summarizing and analyzing existing solutions, we conclude that there is still no solution that provides not only the forecast of a possible drilling accident but also an explanation of why the model made such a forecast.

\subsection{Methods for black-box model interpretability}
Since the drilling engineer requires the model's explanation at a particular prediction, we focus on local interpretability \citep{lipton2018mythos,ribeiro2016should}. Local interpretability aims to understand results for the specific input instead of understanding the entire logic of a trained model including all decision paths, which is in turn called global interpretability. In the following paragraphs, we briefly describe the state-of-the-art solutions, while the complete comparison of the presented methods can be found in papers \citep{elshawi2021interpretability, carvalho2019machine}. The summary of the described approaches is also shown in Table \ref{tab:comparision_2}. In the following paragraphs we will describe advantages and disadvantages of considered approaches and possibility for them to be applied in our study.

\begin{table}[!ht]
\centering
\caption{Comparison of different state-of-art interpretability methods}
\label{tab:comparision_2}
\resizebox{\textwidth}{!}{%
\begin{tabular}{|c|c|c|c|c|c|}
\hline
\textbf{№} &
  \textbf{Group name} &
  \textbf{Methods} &
  \textbf{Idea} &
  \textbf{Drawbacks} &
  \textbf{Papers} \\ \hline
\multirow{3}{*}{1} &
  \multirow{3}{*}{\begin{tabular}[c]{@{}c@{}}Visual \\ investigation\end{tabular}} &
  ICE &
  \begin{tabular}[c]{@{}c@{}}Analyse how the instance's prediction \\ changes when a feature changes \\ using 2D plot\end{tabular} &
  \multirow{3}{*}{\begin{tabular}[c]{@{}c@{}}Hidden logic behind\\  features' relationships \\ can be lost\end{tabular}} &
  \citep{nohara2015interpreting} \\ \cline{3-4} \cline{6-6} 
 &
   &
  \begin{tabular}[c]{@{}c@{}}PCA\\ TSNE\end{tabular} &
  \multirow{2}{*}{\begin{tabular}[c]{@{}c@{}}Display features in 2D or 3D \\ and find similarities between instances\end{tabular}} &
   &
  \citep{wold1987principal, belkina2019automated} \\ \cline{3-3} \cline{6-6} 
 &
   &
  \begin{tabular}[c]{@{}c@{}}Nonlinear \\ dimensionality\\  reduction\end{tabular} &
   &
   &
  \citep{vellido2012making} \\ \hline
2 &
  \begin{tabular}[c]{@{}c@{}}Similarity \\ learning\end{tabular} &
  Analogues search &
  \begin{tabular}[c]{@{}c@{}}Measure distance between \\ train and test instances\end{tabular} &
  \begin{tabular}[c]{@{}c@{}}Does not explain which particular\\  feature influences the forecast\end{tabular} &
  \citep{vellido2012making} \\ \hline
\multirow{4}{*}{3} &
  \multirow{4}{*}{\begin{tabular}[c]{@{}c@{}}Model\\  agnostic \\ models\end{tabular}} &
  LIME &
  \begin{tabular}[c]{@{}c@{}}Explain the black-box model locally \\ by training an interpretable model on \\ perturbed samples and the corresponding \\ black-box prediction.\\ The interpretable model is weighted by \\ the proximity of perturbed instances \\ to the instance of interest.\end{tabular} &
  \begin{tabular}[c]{@{}c@{}}The neighbourhood proximity \\can be improperly defined\end{tabular} &
  \citep{ribeiro2016should, visani2020statistical} \\ \cline{3-6} 
 &
   &
  Anchors &
  \begin{tabular}[c]{@{}c@{}}Explain individual predictions by finding \\ a decision rule that “anchors” the forecast sufficiently. \\ A rule anchors a prediction if changes\\  in other feature values do not affect the prediction.\end{tabular} &
  \begin{tabular}[c]{@{}c@{}}Many scenarios require discretization \\ of features values to obtain rule anchors \\ and get decision boundaries for them\end{tabular} &
  \citep{ribeiro2018anchors, pastor2019explaining} \\ \cline{3-6} 
 &
   &
  SHAP &
  \begin{tabular}[c]{@{}c@{}}Calculate the Shapley value for \\ all possible sets of features \\ for input instance.\end{tabular} &
  Computational time &
  \citep{vstrumbelj2014explaining, guidotti2018survey} \\ \cline{3-6} 
 &
   &
  LORE &
  \begin{tabular}[c]{@{}c@{}}Build a decision tree\\  based on generated samples that are \\ as close as possible to the input instance\\  but correspond to different classes\end{tabular} &
  \begin{tabular}[c]{@{}c@{}}Does not guarantee that it \\ will be able to find a sample \\ from different class\end{tabular} &
  \citep{guidotti2018local, pastor2019explaining} \\ \hline
4 &
  \begin{tabular}[c]{@{}c@{}}Deep \\ learning \\ models\end{tabular} &
  \begin{tabular}[c]{@{}c@{}}Multi-head \\ attention\end{tabular} &
  \begin{tabular}[c]{@{}c@{}}Train attention functions that are learned\\  to understand which input should attend to\end{tabular} &
  \begin{tabular}[c]{@{}c@{}}Requires to use a neural network \\ as the predictive model\end{tabular} &
  \citep{vskrlj2020feature, xiao2017attentional, agarwal2021neural} \\ \hline
\end{tabular}%
}
\end{table}

The first group of methods corresponds to the visual investigation of possible insights of model behavior. This group of methods includes methods both for dimensionality reduction of the feature space and for individual conditional expectation (ICE) \citep{nohara2015interpreting}. In the case of dimensionality reduction, it implies reducing the number of features by feature engineering or such techniques as principal component analysis (PCA) \citep{wold1987principal}, t-distributed stochastic neighbor embedding (TSNE) \citep{belkina2019automated}, nonlinear dimensionality reduction \citep{vellido2012making} techniques. Displaying the features in 2D or 3D, it is possible to notice the similarities between the training sample examples and the input data. In the case of ICE, one may build the plots that show how the prediction for the particular instance changes when a feature value changes. In both dimensionality reduction and ICE cases, the disadvantage is that there might be high dimensional feature spaces where the relationship between features is impossible to understand only by plots. Thus, the hidden logic behind features' relationships can be lost. Since Bag-of-features representation of telemetry logs produce more than $2000$ features, analysis of such pairwise dependencies among features might be inefficient.

The second group of solutions is similar to the first one and aimed at measuring the similarity value of point within the training set at each required time moment \citep{vellido2012making}. Different metrics can measure similarity, for example, general Euclidean distance and its modifications. Unlike the previous group of methods, the current approach does not require reducing the feature space. However, the method's drawback is its dependence on the samples used during the model training. Moreover, the similarity approach will select only similar cases in the training set and not explain which particular feature influences the forecast. 

Another group of interpretability techniques includes methods that are called model agnostic, which can be used for any machine learning model to interpret its behavior. In general, the interpretation is carried out by perturbing the features of the input instance and obtaining the output values of the black-box model for them. This group includes such approaches as local interpretable model-agnostic explanation (LIME) \citep{ribeiro2016should}, Anchors \citep{ribeiro2018anchors}, Shapley additive explanations (SHAP) \citep{vstrumbelj2014explaining}, and local
rule-based explanations (LORE) \citep{guidotti2018local}. The ideas and drawbacks of the particular method are presented in Table \ref{tab:comparision_2}. Compared to the previous group of methods, techniques considered in the current group automatically analyse all possible feature combinations and return the particular feature that influences the forecast more. Thus, current methods can be applied during the current study.

The last group of methods includes the interpretation of neural network (NN) models, which are widely spread in the industry \citep{zhang2018visual, zhang2018opening, liang2021explaining}. This group of methods is based on the interpretation of features extracted by the network on different layers and weights assigned to each feature. Most of the latest methods corresponding to this group are based on the attention mechanism, which description can be found in paper \citep{vaswani2017attention}. Attention mainly gives importance to some input states of the model, which have more contextual relation. So, generally, the weights for the inputs of attention functions are learned to understand which input it should attend to. There are examples presented in papers  \citep{vskrlj2020feature, xiao2017attentional, agarwal2021neural} where it can be used for highlighting the most important features present in the dataset. In our case, the described solution can be used to benchmark the quality of other interpretability techniques since it is the state-of-the-art solution in the industry.

\subsection{Methods for assessing the quality of interpretability}

Like with the interpretability definition, there is also no clear consensus on how to assess its quality \citep{molnar2019}. Nowadays, most researchers use one of the frameworks proposed in the article \citep{doshi2017towards}. The following three paragraphs describe each of the frameworks and the possibility of using each of the them in our case, while the description of the methodology used during the current study is presented in Section \ref{sec:validation}. 

The first framework involves conducting experiments where specialists in a specific area explain the particular feature behavior. Obtaining the specialist's explanation, one may use it as a ground truth value for the explanation provided by the black-box model. Researchers can calculate different quality metrics by comparing the results provided by the model and specialists. In the case of our problem, the current approach implies analysis of telemetry logs and provision of explanation in each time moment for each element in the dataset. In this case, the drilling engineer should specify telemetry parameters in which he or she may see some anomaly patterns. Usage of such an approach is time expensive but guarantees the consistency of interpretation with engineers' knowledge.

Another framework also includes experiments that are carried out with people. However, in this case, the model behavior should be evaluated not with domain experts but with non-specialists. Such evaluation is used when experiments with the target community are challenging or expensive, and the explanation should be provided for some well-known problem (e.g., dogs and cats pictures classification problem). In that case, a user should choose the best explanation from the several proposed ones. Next, similar to the first framework having the most common answers, a researcher may assess the quality of the obtained interpretability. However, since drilling of oil and gas wells is a highly complex problem, evaluation of the model explanation by a non-specialist is impossible, and thus, the current approach can not be applied.  

The last framework is based on function-level evaluation and does not require human opinion. Instead, it uses some formal definition of interpretability as a proxy to assess the quality of the model's explanation. According to the \cite{molnar2022}, the current framework is mostly used when the base classifier model and approach are well-known and have already been evaluated by humans. For example, it might be known that most people understand decision trees. In this case, a proxy for explanation quality may be the depth of the tree. In our case, the current approach can not be applied, since it requires to create some proxy for Bag-of-features model. To our knowledge, there are no proxies consistent with engineers' opinions.

\section{Methodology}
\label{sec:methodology}
\subsection{Overview of Bag-of-features model and its quality}
\label{sec:data_overview}
\begin{figure}[!ht]
    \centering
    \includegraphics[width = 0.9\linewidth]{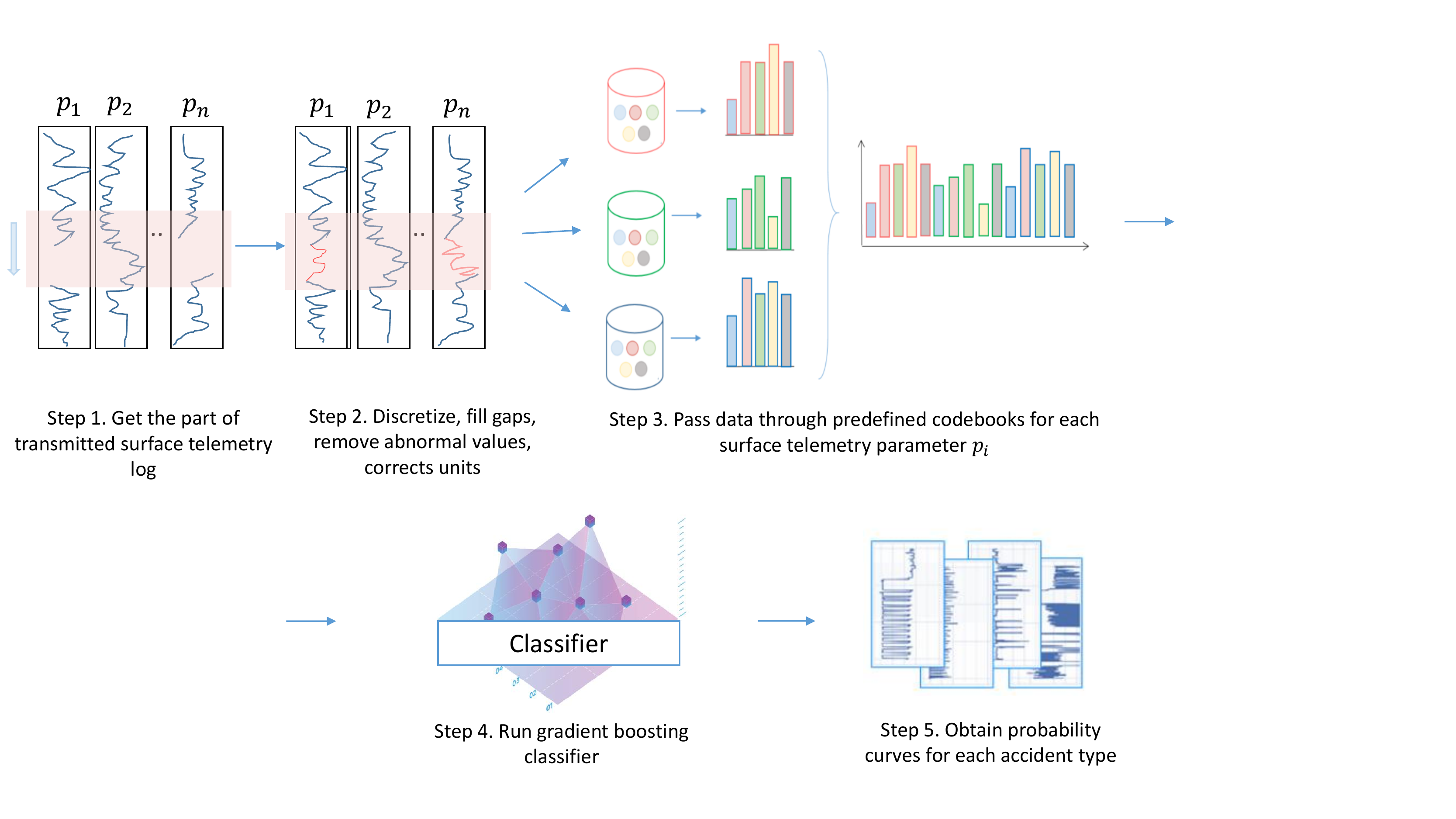}
    \caption{General scheme of Bag-of-features model, used for drilling accident forecasting problem solving \citep{gurina2022forecasting, gurina2021machine}.} 
    \label{fig:hist_gen}
\end{figure}

To interpret the model behavior, one should understand the structure of the input data and the feature engineering process required for the predictive model. This section provides a brief overview of the Bag-of-features model, used for forecasting drilling accidents in real-time and based on the Bag-of-features approach. As was mentioned in Section \ref{sec:intro}, the full description of the Bag-of-features model can be found in papers \citep{gurina2022forecasting, gurina2021machine}, while the general scheme of the model is presented in Figure \ref{fig:hist_gen}. 

For input data, the model requires the one-hour-length segment of the telemetry log containing $12$ parameters, also required for the drilling engineer to forecast accidents. Such parameters and their abbreviation that are mostly used in Wellsite Information Transfer Standard Markup Language (WITSML) \citep{kirkman2003wellsite, dict_mnemonic} are the following:

\begin{itemize}
  \item Hookload average (HKLA)
  \item Weight on the bit (WOB)
  \item Block position (BPOS)
  \item Depth of the Bit True Measured (DBTM) and Hole True Measured Depth (DMEA)
  \item Torque avarage (TQA) and Rotations Per Minute Average (RPMA)
  \item Standpipe Pressure (SPPA)
  \item Mudflow (MFIA, MFOA) in and out of the pipe avarage
  \item Total volume in tanks (TVT)
  \item Gas content avarage (GASA). 
\end{itemize}

Next, the input segment is passed through the discretization procedure, where the data is sampled with the 10-second step. At the same time, anomaly and missing values are replaced with valid ones. In the case of anomaly values, if the considered value were higher or lower than the predefined limits for the particular parameter, the value was replaced with the previous valid value. The values of predefined limits can be found in paper \citep{gurina2022forecasting}. Similar to anomaly filtration, if there are some missing values, the model propagate the last valid observation forward.

Finally, using the sliding window technique for each telemetry parameter, the input segment is divided into small segments named $\tau$-parts. By clustering the obtained $\tau$-parts with the previously trained clustering models (codebooks), one can build a histogram of cluster labels for the one-hour-length input segment of the telemetry log. Stacking the histogram for all telemetry parameters together, one can obtain the final feature vector required for the gradient boosting model to forecast the accident of the particular type.  

The gradient boosting classifier used as an approximation function $\hat{F}(x)$ for the log-odds of the positive class of the binary variable $y$ (whether the corresponding intervals contain some accidents and pre-accident patterns or not), where the feature vector $x$, generated with Bag-of-features approach. The full description of gradient boosting method can be found in paper \citep{Chen}. The gradient boosting classifier is a logistic transformation of the weighted sum of log-odds returned by decision tree estimators $h$:

\begin{align}
    \hat{F}(x) = \sum_{i=0}^{M}(\gamma_{i}*h_{i}(x)) + const
\end{align}

\begin{align}
    P(y=1) = \frac{1}{1+\texttt{exp}\left(-\hat{F}(x)\right)}
\end{align}

where $M$ is a number of decision trees in the classifier, and $\gamma_{i}$ is a weight coefficient for the corresponding tree decision. The corresponding approximation is achieved through minimizing the expectation $E_{x, y}$ between the $\hat{F}(x)$ and the true function $F(x)$ by introducing loss function $L(y, F(x))$:
\begin{align}
    \hat{F}(x) = argmin(E_{x, y}(L(y, F(x))))
\end{align}

In our case, we use gradient boosting classifier with logistic loss function ($L$), 250 estimators ($M$). In addition, we set the following parameters for training and decision tree estimators: learning rate equals 0.05, max depth of each tree is 10, while the subsample and colsample by tree correspond to 90\%, the weight of the positive class is increased five times. The description of the corresponding parameters is presented in \citep{XGB, Chen}.

To assess the Bag-of-features model's quality, we propose using the Receiver operating characteristic (ROC) curve and calculating the area under it (ROC AUC metric). The description of the proposed metric can be found in paper \citep{Foundation_of_ML}. We assume that the accident is forecasted correctly if the model alarm happens inside the reference region and the type is identified correctly. Otherwise, the model warning is counted as a false alarm. Using a $5$-fold cross-validation procedure, where each well is included in the test set once per cross-validation, one can obtain probabilities for the whole set of wells and build the final metric plot using a micro-averaging aggregation strategy. 

For training and validation of the Bag-of-features model we use fragments of real-time telemetry data with the drilling parameters, presented above. As it was shown in papers \citep{gurina2022application, gurina2022forecasting}, to solve accident drilling forecasting problem we collected the database of drilling accidents and pre-accident lessons that contain the exact date-time or depth of these events. Next, we match the obtained event information with the telemetry data for the particular well. In doing this, we extract more than 1000 drilling accident predecessors segments for more than 100 real accidents and about 2500 normal drilling intervals and use them as a training set for the model. 
The distribution of different accident types and the number of predecessors and accidents are presented in Table \ref{tab:dist}.

\begin{table}[!ht]
\centering
\caption{The distribution of different types of accidents and the number of predecessors and accident intervals in the dataset used for the Bag-of-features training}
\label{tab:dist}
\resizebox{\textwidth}{!}{%
\begin{tabular}{|cc|c|c|}
\hline
\multicolumn{1}{|c|}{№} & Accident type & Number of   predecessor intervals & Number of accident intervals \\ \hline
\multicolumn{1}{|c|}{1} & Stuck     & 86   & 69  \\ \hline
\multicolumn{1}{|c|}{2} & Mudloss   & 6    & 30  \\ \hline
\multicolumn{1}{|c|}{3} & Kick flow & 901  & 10  \\ \hline
\multicolumn{1}{|c|}{4} & Washout   & 23   & 20  \\ \hline
\multicolumn{2}{|c|}{Total}         & 1016 & 129 \\ \hline
\end{tabular}%
}
\end{table}

\subsection{Methodology for Bag-of-features interpretation}
For the Bag-of-features representation, the explanatory model should analyze the feature vector of $2400$ length in each time moment since each of the codebooks is represented by the $200$ clusters. The particular item in the feature vector is the amount of $\tau$ segments corresponding to the particular cluster, which in turn stands for a short time series pattern represented by $\tau$ segments obtained during the training of the codebooks. By analyzing the frequency of occurrence of such patterns over one hour, the model can predict an accident. 

The logic behind such a feature engineering procedure can be explained with the following example. The occurrence of several rapid increases of TQA while drilling for a certain period might be considered as predecessor signs for future stuck accidents. In this case, the $\tau$ segment with torque increase represents such short time-series pattern. Thus, the selection of $\tau$ segments that are important for the model in a particular situation may give the engineer an explanation of the model intelligence. 

\begin{figure}[!ht]
    \centering
    \includegraphics[width = 0.9\linewidth]{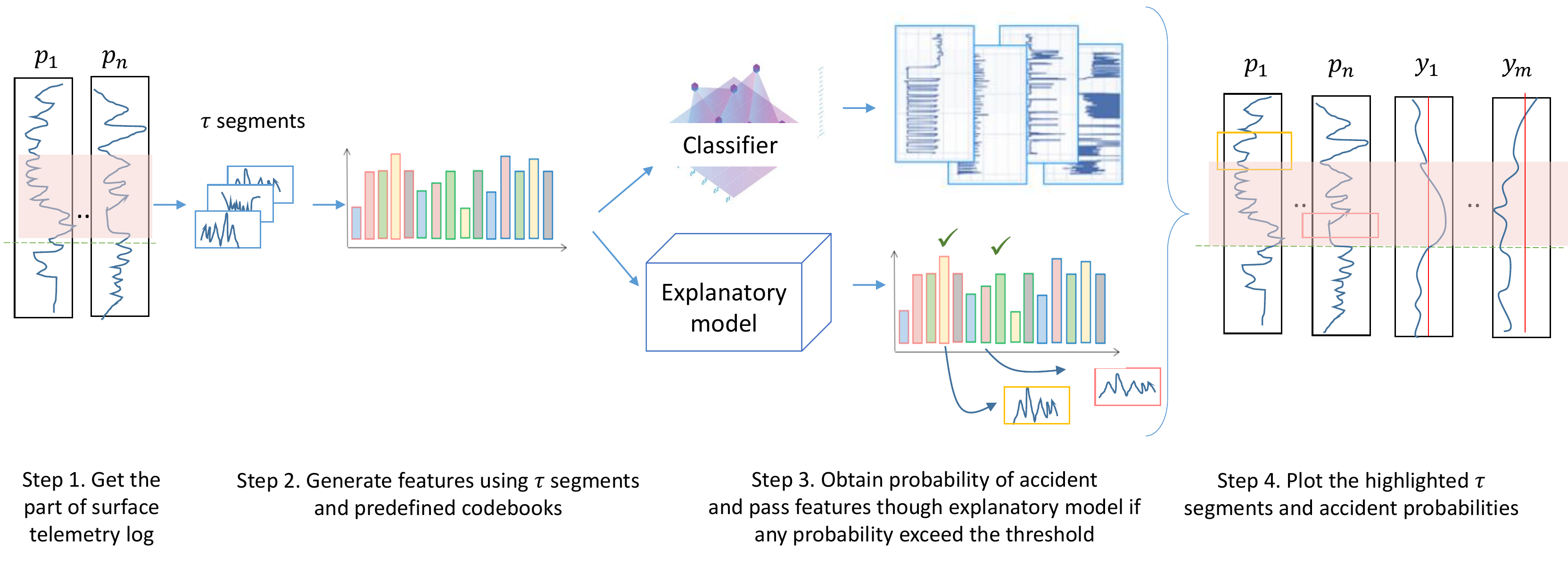}
    \caption{General scheme for interpretation of Bag-of-features model, used for drilling accident forecasting problem \citep{gurina2022forecasting, gurina2021machine}.} 
    \label{fig:sharp_gen}
\end{figure}

To obtain the top-$K$ most important features with the explanatory model, we should highlight all the $\tau$ segments that stand for all top-$K$ clusters. The selection of such $\tau$ segments would be the visual representation of the rationale for why a model's prediction was made. Since the $\tau$ segments are generated with the sliding window technique with a one-minute step, they usually intersect in a certain area of the time series. The scheme of the proposed methodology is presented in Figure \ref{fig:sharp_gen}.

To obtain the top-$K$ most important items from the feature vector, we decided to highlight only those segments whose importance value is higher than some percentage $M$ from the maximum importance value for the considered feature vector. The hyperparameter $M$ percentage should be chosen separately during the optimization stage for each interpretation algorithm. The values of hyperparameter $M$ obtained for each interpretation algorithm are presented in Section \ref{sec:results}. 

\begin{figure}[!ht]
    \centering
    \includegraphics[width = 0.9\linewidth]{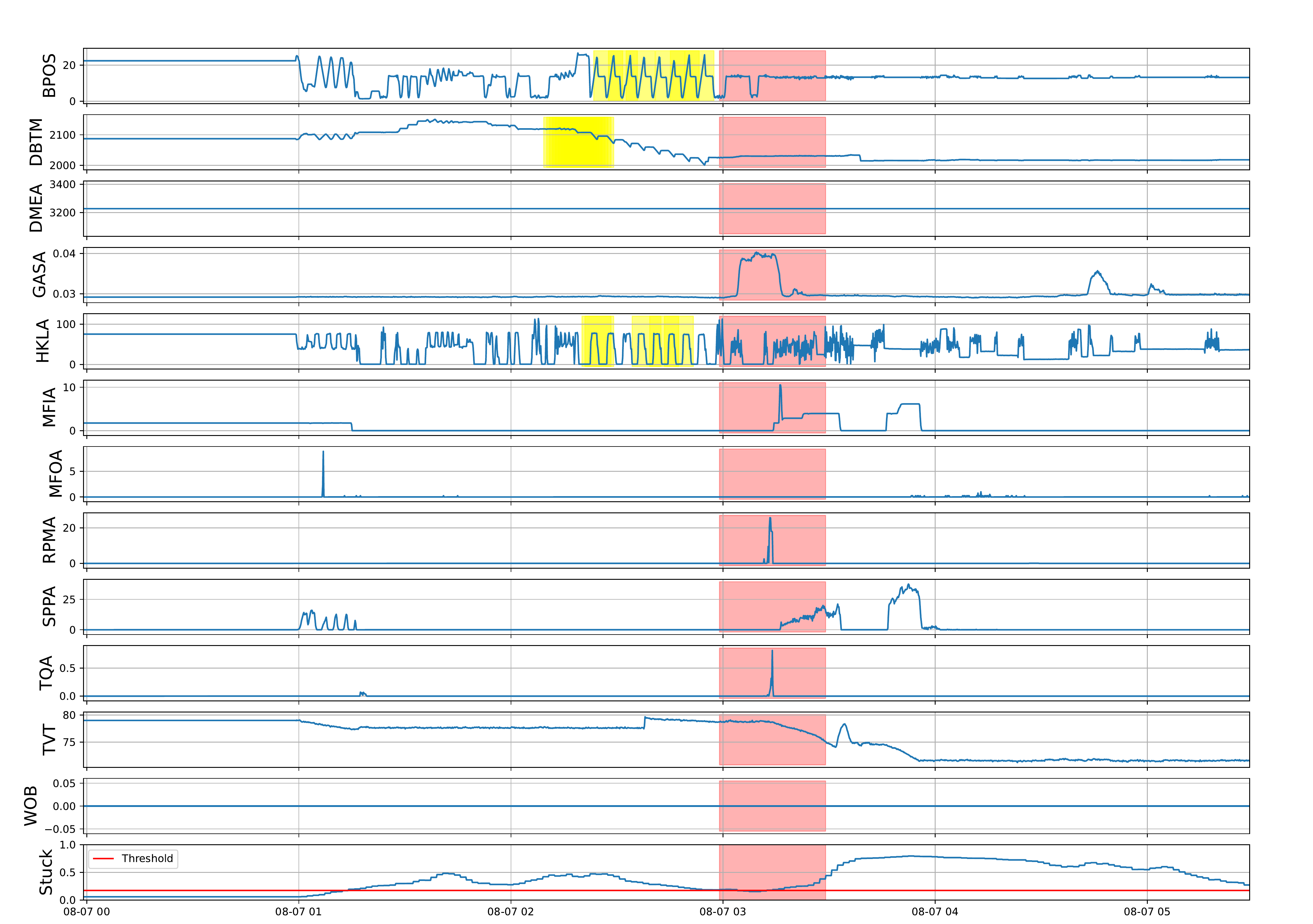}
    \caption{Example of provided interpretation for the stuck accident. The yellow area stands for the highlighted $\tau$ segments provided using the explanatory model. The red zone corresponds to the region selected by the drilling engineer, where the actual stuck accident happened. The last row shows the probability of a stuck accident, obtained using the Bag-of-feature model.} 
    \label{fig:inter_example}
\end{figure}

Summing up, according to the proposed methodology, when the probability of a particular accident is higher than the chosen threshold, we highlight $\tau$ segments corresponding to the top-$K$ significant clusters and provide the probability of an accident. An example of the obtained interpretability is presented in Figure \ref{fig:inter_example}. The yellow area stands for the highlighted $\tau$ segments provided using one of the explanatory models. The red zone corresponds to the region selected by the drilling engineer, where the actual stuck accident happened. The last row shows the probability of a stuck accident, obtained using the Bag-of-feature model.

\subsection{Interpretation methods used in the current study}
\label{sec:set_up}
In Section \ref{sec:intro}, we considered latest methods for the interpretability of black-box models. In application to our study, we decided to use the SHAP method since it provides a marginal distribution of features contributions across all combinations and shows better results than other model agnostic approaches in similar problems \citep{ramon2020comparison, man2021best}.

To benchmark the interpretability obtained using the SHAP method with the state-of-the-art methods, we use a neural network with a multi-head attention mechanism, described in Section \ref{sec:state-of-art} and Table \ref{tab:comparision_2}. We replace the gradient boosting classifier of the Bag-of-features model with a fully connected network that has a multi-head attention mechanism (FCMH). It uses a Bag-of-features representation of telemetry logs as features and predicts accident probabilities for each time step. Like SHAP, the model also returns the vector, showing the importance of a particular item presented in the input feature vector (feature importance vector). The FCMH model has two main units: the classifier and multi-head attention layer with three independent linear layers inside. The classifier unit contains two fully connected linear layers with $2400$ and $64$ input neurons and a dropout layer (p=0.05). During training, we used SGD optimizer and cross-entropy as loss functions.

We also decided to compare the performance of SHAP and FCMH methods with random and baseline models. For the random model, we sample 10 vectors with random feature importance for each moment, where the Bag-of-features model exceeds the threshold, and calculate metric values using a micro-averaging aggregation scheme. In the case of the baseline model, we assume that all features are equally important, and thus we select all $\tau$ segments. 

\subsection{Validation procedure}
\label{sec:validation}
Considering the methods for assessing the quality of interpretability presented in Section \ref{sec:methodology}, we decided to use the combination of the functional evaluation approach together with explanations provided by the expert. We assume that the model's interpretability is good if the highlighted $\tau$ segments, chosen using a particular method, correspond to the regions highlighted by the drilling expert. In particular, we suppose that the $\tau$ segment was highlighted correctly if it intersects with the expert's reference for the particular telemetry parameter. Otherwise, the provided model explanation was counted as a false one. On average, the drilling engineer reference contains two or three main telemetry parameters for each accident type, which behavior is selected as an abnormal one. By calculating the precision and recall metrics (the definition can be found in paper \citep{buckland1994relationship}) using a particular vector of thresholds and micro-averaging aggregation over the considered accident types, one can assess the quality of the provided interpretation. 

Since the Bag-of-features model can find the hidden relationships between parameters, we also decided to assess interpretation quality with some concessions. Similar to the previous case, we assumed that the $\tau$ segment was highlighted correctly if it intersects with the expert's reference, but the list of possible highlighted telemetry parameters was extended according to their relevance for the particular accident type (Table \ref{tab:MWD_params}).

\begin{table}[!ht]
\centering
\caption{Set of telemetry parameters that can be highlighted for each accident type.}
\label{tab:MWD_params}
\resizebox{0.85\textwidth}{!}{%
\begin{tabular}{|c|c|c|}
\hline
№ & Accident type & Set of telemetry parameters \\ \hline
1 & Kick flow & GASA, TVT, MFIA, MFOA, DBTM, DMEA \\ \hline
2 & Mudloss & TVT, SPPA, MFIA, MFOA, DBTM, DMEA \\ \hline
3 & Stuck & HKLA, BPOS, WOB, TQA, RPMA, DBTM, DMEA \\ \hline
4 & Washout of drill pipe & TVT, SPPA, MFIA, MFOA, TQA, DBTM, DMEA \\ \hline
\end{tabular}%
}
\end{table}

To obtain the expert reference regions, we selected five wells with accident cases for the kick flow, stuck, mudloss, and washout as the most common accidents in the industry. For each case, we ask the drilling expert to analyze telemetry logs and special telemetry features that correspond to why the particular accident occurred. For all cases, the expert uses only the parameters described in Section \ref{sec:data_overview} and highlights the interval, which shows some anomaly behavior. For example, for the stuck case, presented in Figure \ref{fig:inter_example}, the drilling engineer selected HKLA and BPOS features as the most valid ones for that accident.

\section{Results and discussions}
\label{sec:results}
\subsection{General interpretability quality of Bag-of-features model}

\begin{figure}[!ht]
\centering
\includegraphics[width = 0.65\linewidth]{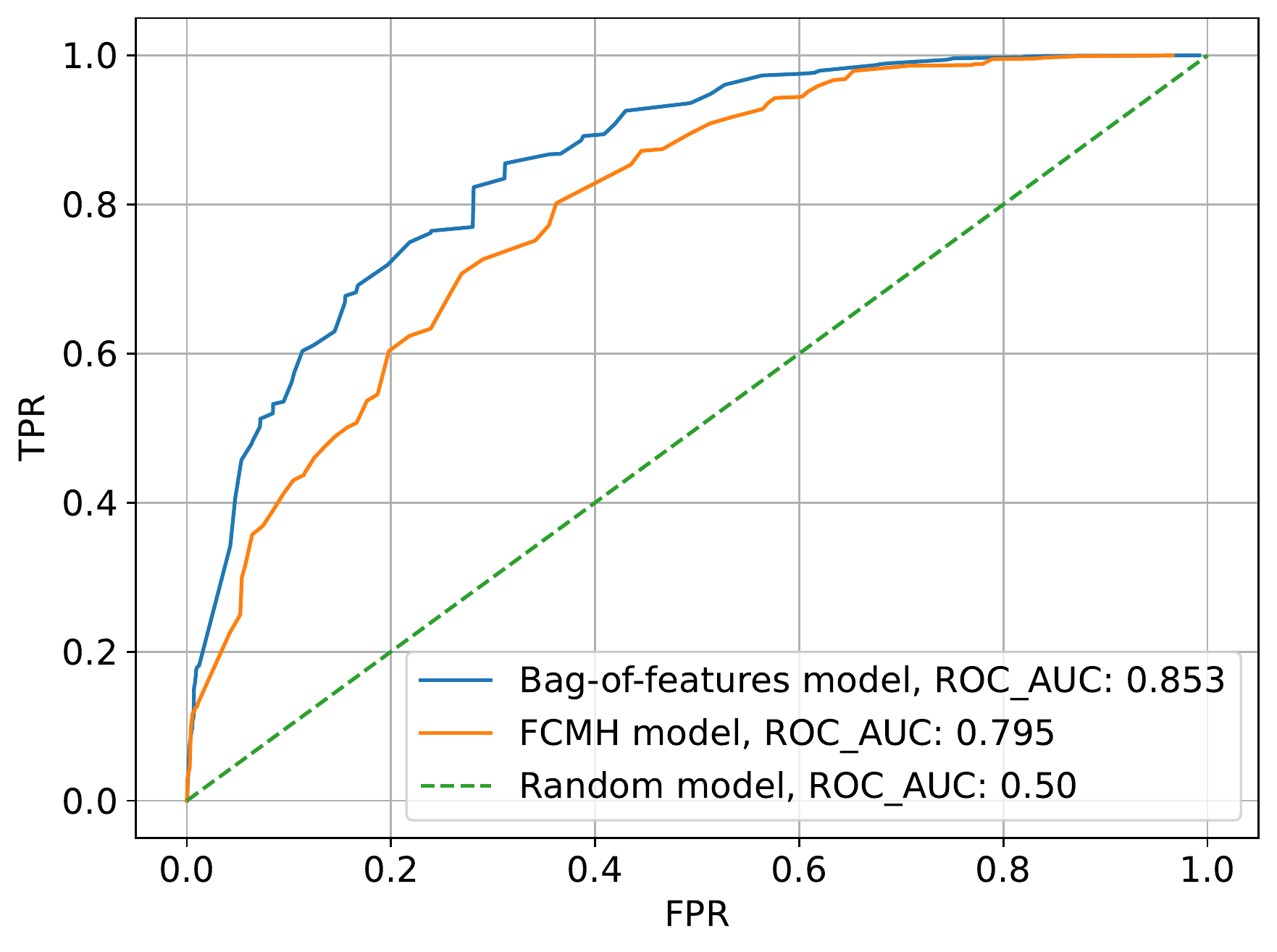} 
\caption{Accident forecasting quality metrics for the Bag-of-features, FCMH, and Random models. The FCMH model has a similar quality to the Bag-of-features model.}
\label{fig:quality_BOF}
\end{figure}

According to the experimental setup described in Section \ref{sec:set_up}, during this study, we compare the SHAP and FCHM interpretability techniques together with random and baseline models. First, to compare interpretability techniques for the FCMH model, we achieved a similar quality to the Bag-of-features model. The corresponding metric plots for Bag-of-feature and FCMN models are shown in Figure \ref{fig:quality_BOF}.

Based on the methodology described in Section \ref{sec:validation}, the interpretation quality was estimated using the precision and recall metrics obtained for each accident type's particular probability threshold. We decided to use different thresholds for each accident type to achieve the minimum number of false alarms having the maximum percentage of covered cases. For the Bag-of-features model, we choose the threshold corresponding to 70\% of covered cases with one false alarm per day for the kick flows accidents. For the stuck and washout accidents, we choose the threshold corresponding to 60\% of covered cases with correspondingly four and three false alarms per day. For the mudloss accident, the threshold stands for 55\% of covered cases and two false alarms per day. At each moment where the probability of an accident provided by the Bag-of-features model exceeds the corresponding threshold value, one obtains the feature importance using SHAP, FCMH, baseline, and random algorithms. It was mentioned in Section \ref{sec:methodology} that each interpretation algorithm used different value of hyperparameter $M$. The interpretation results were obtained for such a value of $M$, which provides the highest value of the metrics for each algorithm. In the case of the SHAP algorithm, we obtain $M$ equal to $20\%$, while in the case of FCMH $M$ value was $24\%$. For baseline and random models, the value of hyperparameter $M$ was set to $30\%$.

The precision and recall metrics for described models are presented in Table \ref{tab:pr_recall_res}. Based on the obtained results, one may notice that the SHAP algorithm provides better results than the FCMH, random, and baseline models for both experiments. Thus, it will be used as an explanatory model for further experiments. The recall metric is high for all algorithms, which means that almost all highlighted $\tau$ segments intersect with expert reference. However, at the same time, the values of precision metric suggest that most of the errors are related to the misinterpretation of which telemetry features influence the most at the current time moment. It means that the Bag-of-features model found and used some additional hidden relationships between telemetry parameters that are not presented in Table \ref{tab:MWD_params}. The found patterns are meaningful in terms of understanding the drilling operations and drilling behavior in general. In contrast, the reference provided by the drilling engineer only indicated the regions of some anomaly behavior and not the patterns related to the particular drilling operation. 

\begin{table}[!ht]
\centering
\caption{The metric values were obtained for different interpretation algorithms. SHAP algorithm gives twice higher precision metric values than the FCMH, random, and baseline models with similar level of recall metric.}
\label{tab:pr_recall_res}
\resizebox{\textwidth}{!}{%
\begin{tabular}{|c|cc|cc|}
\hline
 &
  \multicolumn{2}{c|}{Metrics on drilling expert reference} &
  \multicolumn{2}{c|}{\begin{tabular}[c]{@{}c@{}}Metrics on drilling expert reference and \\ extended list of telemetry parameters\end{tabular}} \\ \hline
       & \multicolumn{1}{c|}{Presicion} & Recall & \multicolumn{1}{c|}{Precision} & Recall \\ \hline
SHAP  & \multicolumn{1}{c|}{0.1589}    & 0.7025 & \multicolumn{1}{c|}{0.2855}    & 0.6625 \\ \hline
FCMH   & \multicolumn{1}{c|}{0.0727}        & 0.6945     & \multicolumn{1}{c|}{0.1797}        & 0.6435     \\ \hline
Baseline & \multicolumn{1}{c|}{0.0519}    & 1.0 & \multicolumn{1}{c|}{0.1731}    & 1.0 \\ \hline
Random & \multicolumn{1}{c|}{0.0635}    & 0.7289 & \multicolumn{1}{c|}{0.2012}    & 0.678 \\ \hline

\end{tabular}%
}
\end{table}

\subsection{Consistency of explanatory model}
After introducing the general model quality metrics, we conducted several experiments to check the consistency of the best explanatory model. Consistency check can be done using the TSNE technique that, as input, use a pairwise distance matrix between features of highlighted $\tau$ segments and $\tau$ segments used during the codebook training. If segments highlighted by the explanatory model are similar, then the corresponding points should be close. 

\begin{figure}[!ht]
    \centering
    \includegraphics[width = 0.9\linewidth]{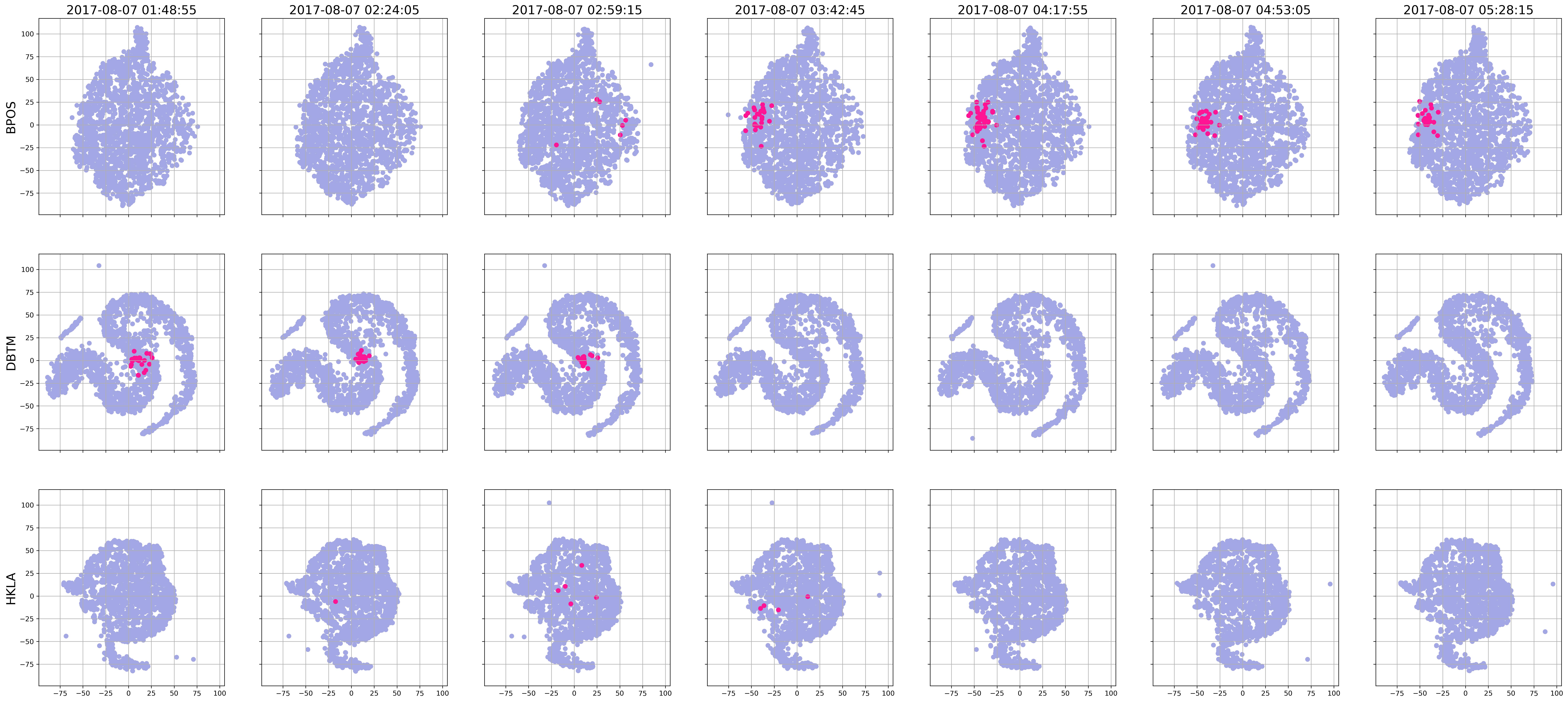}
    \caption{TSNE plot for the neighboring time moments for the stuck accident, shown in Figure \ref{fig:inter_example}. The pink points correspond to the $\tau$ segments highlighted by the explanatory model at each moment, while the purple points represent the $\tau$ segments used during training the corresponding codebook. For those telemetry parameters that are not presented in the plot, there were no highlighted $\tau$ segments. For neighboring time moments, the highlighted $\tau$ segments are located in the same area of the TSNE plot for each telemetry parameter, which means that for the current case, we have a good consistency for the explanatory model.}  
    \label{fig:tsne_exp}
\end{figure}

First, the obtained explanatory model should identify similar tau segments for neighboring time moments since changes in features at neighboring time moments are not so significant. Due to the limited number of plots that can be added to the article, we decided to show the TSNE representation (Figure \ref{fig:tsne_exp}), obtained only for the stuck case, shown in Figure \ref{fig:inter_example}. The TSNE representation is shown for the particular telemetry parameter and seven neighboring time moments where the stuck probability exceeds the threshold. The pink points correspond to the $\tau$ segments highlighted by the explanatory model at each moment, while the purple points represent the $\tau$ segments used during training the corresponding codebook. For those telemetry parameters that are not presented on the plot, there were no highlighted $\tau$ segments.

For neighboring time moments, the highlighted $\tau$ segments are located at the same area of the TSNE plot for each telemetry parameter. It means that we have a good consistency for the current accident case and can further use the proposed interpretability approach. We also checked consistency using that methodology and got similar results for the other cases presented in the holdout dataset.

Second, the obtained explanatory model should identify similar tau segments for accident cases that correspond to the same accident type. Similar to the previous case, we plotted the TSNE representation for each accident type (Appendix \ref{TSNE_for_different_type}), whereby in the pink color, we plot the highlighted $\tau$ segments for all-time moments, where the probability of a particular accident exceed the threshold. By purple color, we plot the TSNE representation of $\tau$ segments used during the training of the corresponding codebook. By yellow color, we represent $\tau$ segments obtained on intervals, that were selected by the drilling engineer as abnormal one for the particular accident case.

The highlighted $\tau$ segments are located in different regions and stand for different telemetry parameters for each accident type. We conclude that the proposed interpretation is consistent with different drilling accidents based on the results. In most cases, the explanatory model's selected $\tau$ segments are located in the same area as $\tau$ segments obtained from anomaly regions highlighted by the drilling engineer. In particular, we conclude that during the forecast of kick flow accidents, the Bag-of-features model mostly looks at changes in the GASA parameter, while the stuck forecast model usually pays attention to changes in BPOS, DBTM and HKLA parameters. In the case of a washout accident, the essential parameters for the Bag-of-features model are MFIA, MFOA, TVT, HKLA, and DBTM features. For mudloss accidents model usually focuses on such features as MFIA, SPPA, and TVT.

The obtained results are consistent with the physics of the drilling process. The drilling engineer also uses the selected parameters for each accident type to forecast the particular accident type. 

\subsection{Discussion and future work}
Minimizing NPT during drilling is essential for the oil and gas industry. Nowadays, drilling engineers' knowledge and machine learning models based on Bag-of-features representation of telemetry logs allow to minimize NPT by forecasting the drilling accidents and detecting pre-accident signs. However, using such machine-learning models requires a high level of trust by the drilling engineer in the system, which is challenging for black-box models. This problem can be alleviated by explaining the decisions made by such models in terms of highlighting the key features they are based on (see sections \ref{sec:intro}, \ref{sec:methodology}). 

The quality of the explanatory model used the SHAP approach surpasses metrics of both baselines and the FCHM model. Moreover, the proposed model is consistent with the drilling process and model performance. However, there is still room for improvements. The quality of the proposed explanatory model should also be tested during real-time operation. In particular, it is interesting to research the predictive model whose interpretation is as close as possible to the drilling engineer reference. Moreover, it will be helpful to provide an explanation not only highlighting $\tau$ segments but also with a general description of patterns presented in a particular cluster. 

\section{Conclusions}
\label{sec:conclusion}
We proposed an approach disclosing how Bag-of-features model that used for drilling accident forecasts can be interpreted. The developed explanatory model based on the SHAP method and allows to highlight parts of telemetry logs that stand for some abnormal behavior. 

The explanatory model can select 70\% of anomaly telemetry segments and choose 15\% of telemetry parameters, similarly to a drilling engineer. The precision metric is higher than the precision metrics for the random, baseline, and state-of-the-art solutions obtained with the same level of recall metric. Metrics demonstrate that the model is of reasonable quality and consistent with the physics of the drilling process. 

Nowadays, both Bag-of-features and explanatory model, based on the SHAP approach, are being tested in real oilfields. To operate, we developed software integrated with the Wellsite Information Transfer Standard Markup Language data server into clients' existing IT infrastructure. All calculations take place in the cloud and, therefore, do not require significant additional computing power on the client-side.

The joint usage of the Bag-of-features predictive model and the explanatory model is to prevent more drilling accidents. This results in less NPT by increasing the trust level to the model predictions and AI technologies applied for drilling in general. 

\section{Statements and Declarations}
\subsection{Funding}
This research did not receive any specific grant from funding agencies in the public, commercial, or not-for-profit sectors.

\subsection{Conflict of interest/Competing interests}
The authors have no conflicts of interest to declare that are relevant to the content
of this article.

\subsection{Availability of data and materials}
Data is not available by the non-disclosure agreements with companies.

\subsection{Code availability}
Code is not available by the non-disclosure agreements with companies.

\bibliographystyle{elsarticle-harv}
\bibliography{Inter_Bag_of_features}

\begin{appendices}
\section{}
\begin{figure}[!ht]
    \centering
    \includegraphics[width = \linewidth]{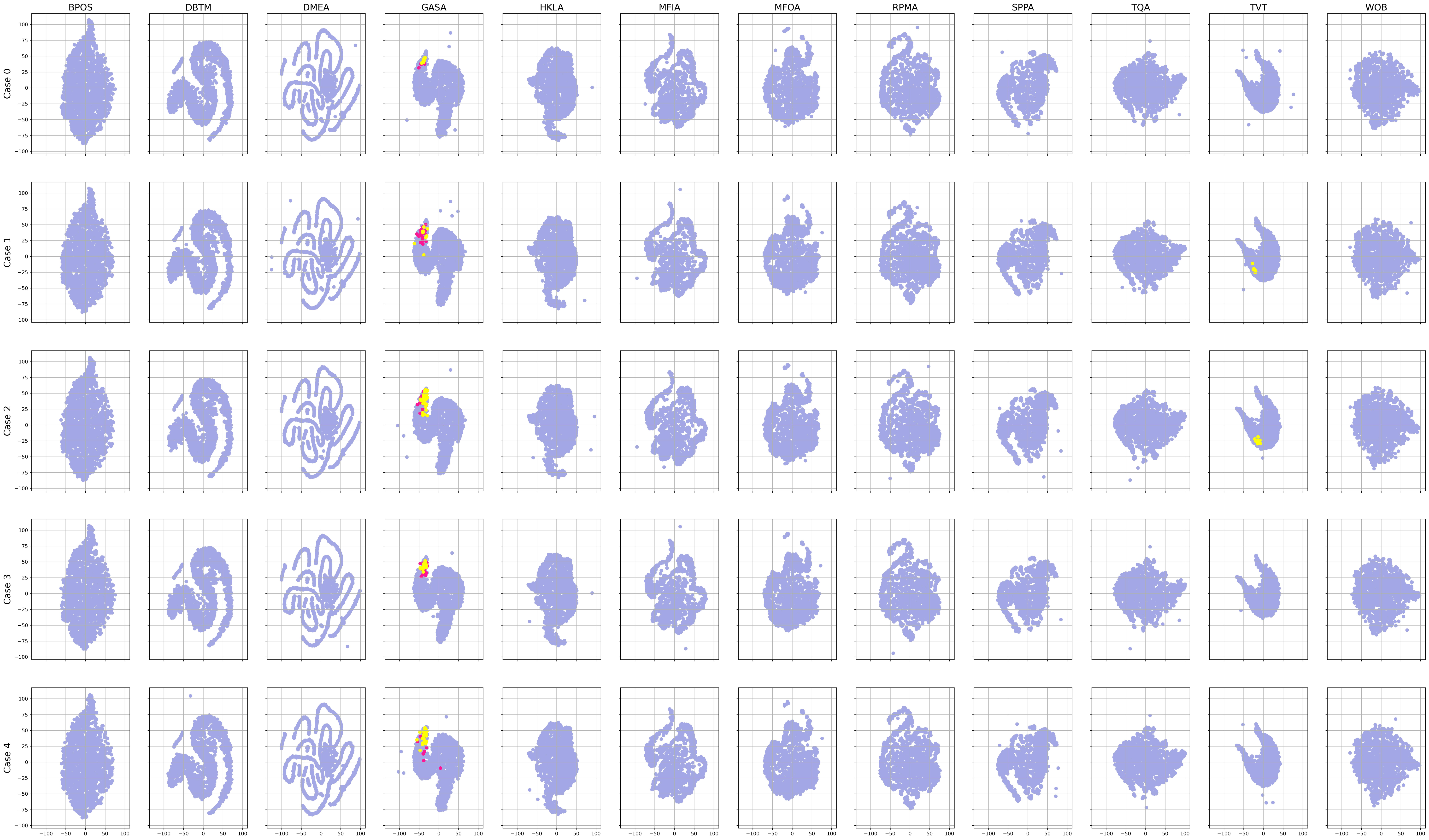}
    \caption{TSNE for kick flow accidents}  
    \label{fig:tsne_fluid_show}
\end{figure}

\begin{figure}[!ht]
    \centering
    \includegraphics[width = \linewidth]{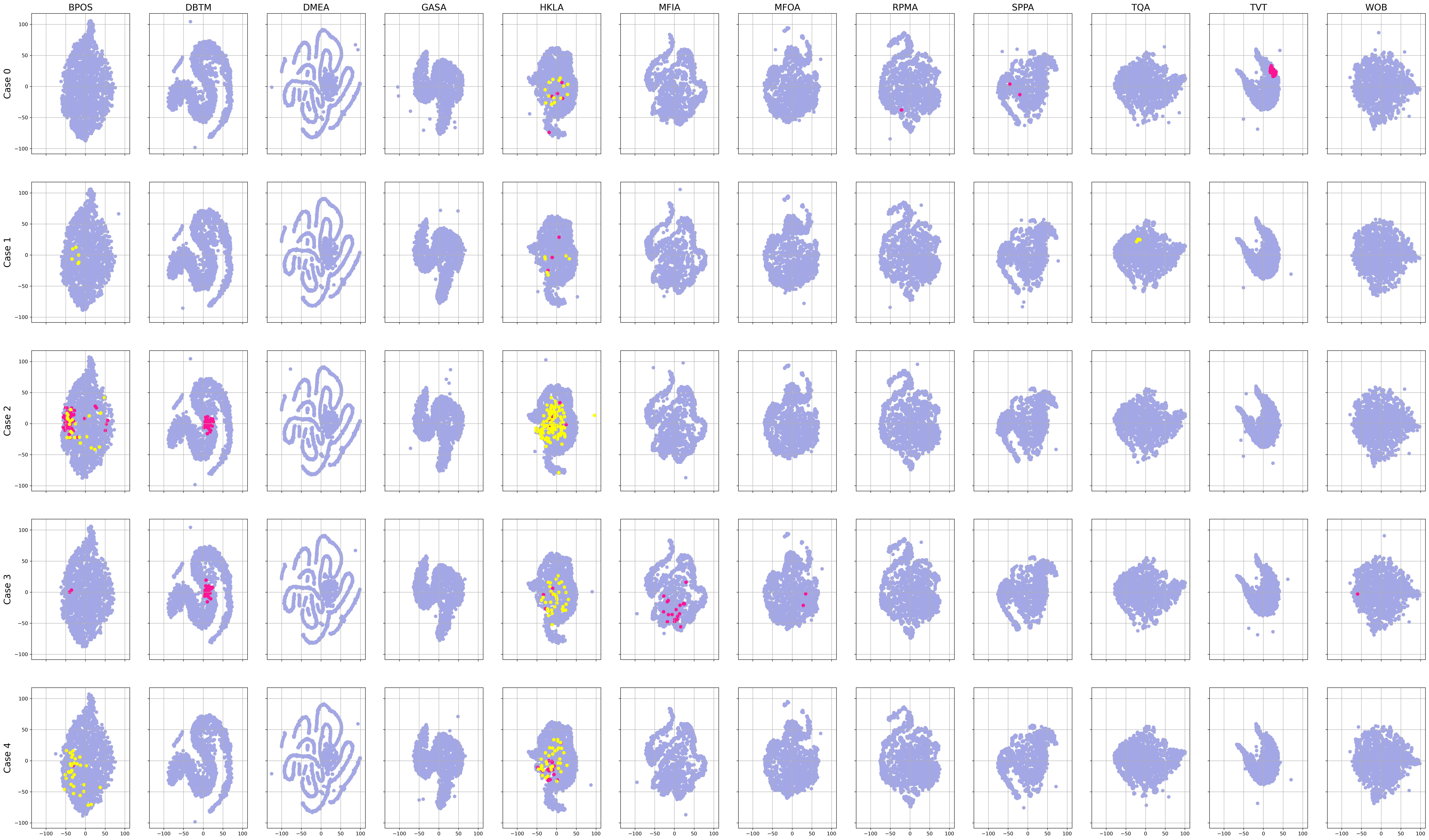}
    \caption{TSNE for stuck accidents}  
    \label{fig:tsne_stuck}
\end{figure}

\begin{figure}[!ht]
    \centering
    \includegraphics[width = \linewidth]{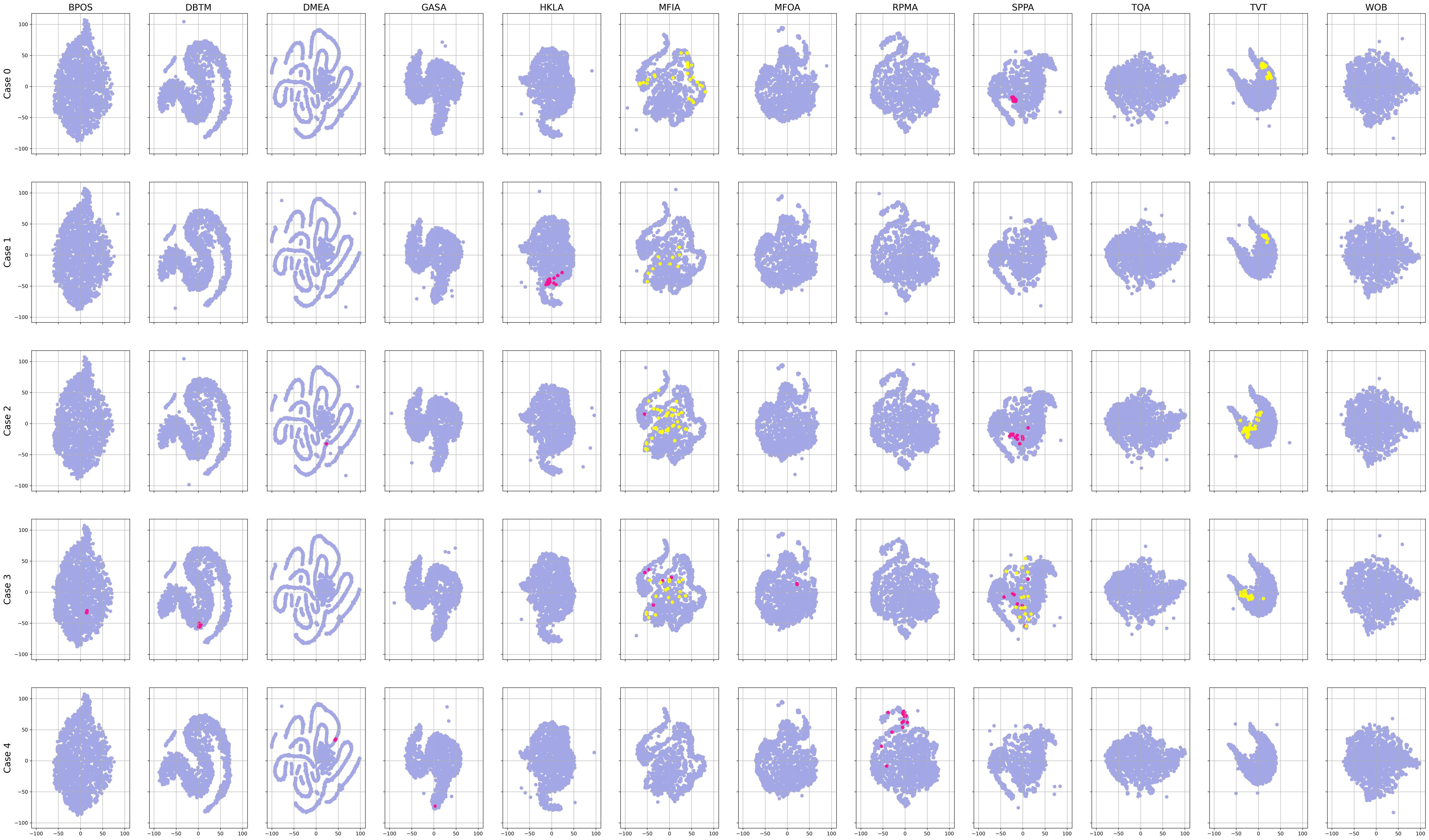}
    \caption{TSNE for mudloss accidents}  
    \label{fig:tsne_mudloss}
\end{figure}

\begin{figure}[!ht]
    \centering
    \includegraphics[width =\linewidth]{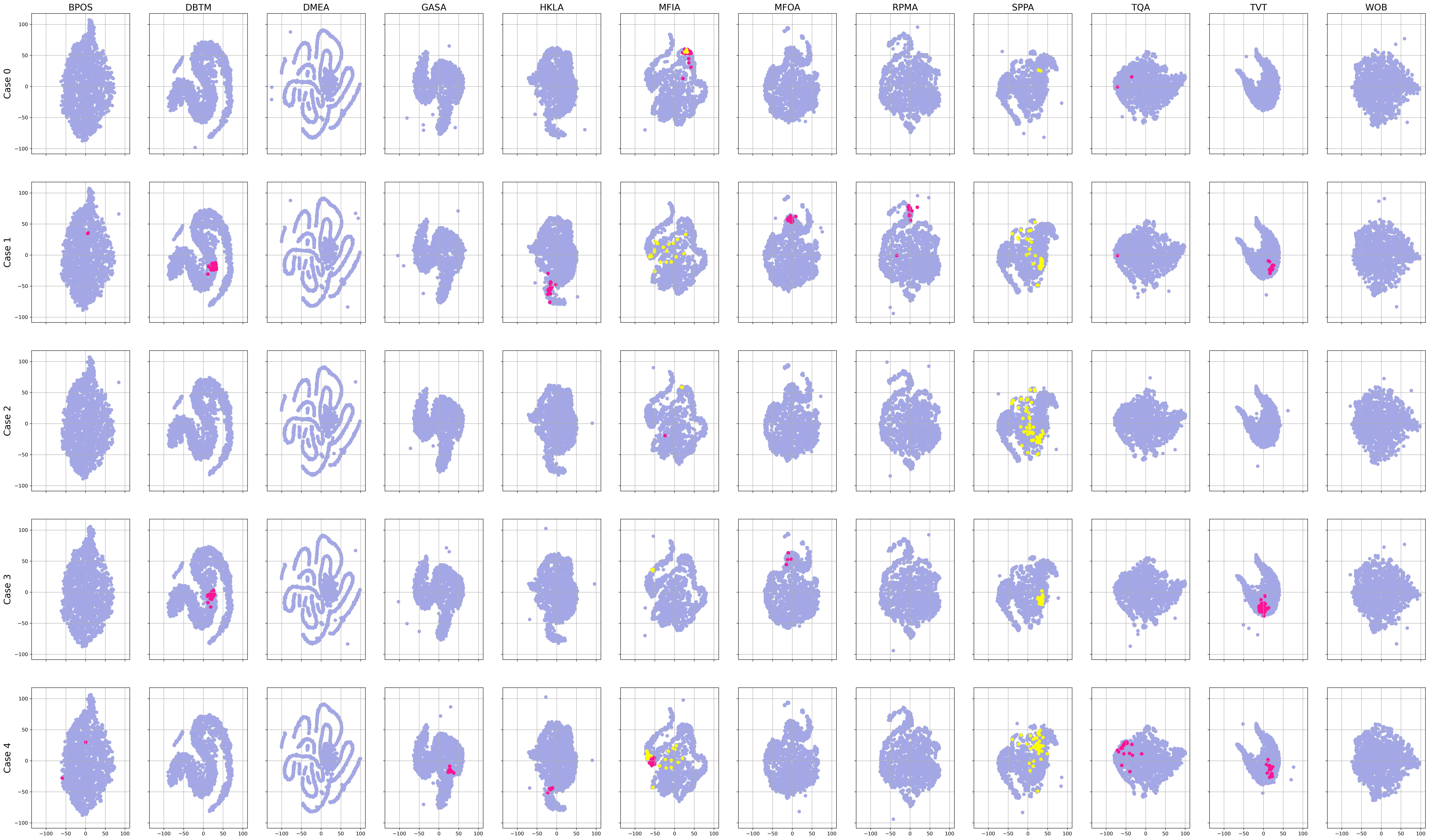}
    \caption{TSNE for washout accidents}  
    \label{fig:tsne_washout}
\end{figure}

In Figures \ref{fig:tsne_fluid_show}-\ref{fig:tsne_washout}, one may see the TSNE plots for different types of accidents, whereby the pink color we plot the highlighted $\tau$ segments by the explanatory model for all time moments, and by purple color, we plot the TSNE representation of $\tau$ segments used during the training of the codebook. By yellow color, we represent $\tau$ segments obtained on intervals, that were selected by the drilling engineer as anomaly one for the particular accident case.

\label{TSNE_for_different_type}
\end{appendices}

\end{document}